\documentclass{llncs}
\usepackage{multirow}
\usepackage{caption}
\usepackage{booktabs}
\usepackage{amsmath}
\usepackage{amssymb}
\usepackage{epsfig}
\usepackage{amsfonts}
\usepackage{longtable}
\usepackage{rotating}
\usepackage{color}
\usepackage{hyperref}
\hypersetup{
    colorlinks=true,
    linkcolor=blue,
    filecolor=magenta,
    urlcolor=blue,
}

\begin{document}
\frontmatter          
\pagenumbering{arabic}

\title{DiamondGAN: Unified Multi-Modal \\
Generative Adversarial Networks \\
for MRI Sequences Synthesis}

\author{Hongwei Li$^{1}$\thanks{equal contribution} \and Johannes C. Paetzold$^{1\star}$ \and Anjany~Sekuboyina$^{1, 2}$ \and Florian~Kofler$^{1}$ \and Jianguo Zhang$^{3, 4}$ \and Jan S. Kirschke$^{2}$ \and Benedikt Wiestler$^{2}$ \and Bjoern Menze$^{1, 5}$}
\institute{1. Dept. of Informatics, Technical University of Munich, Germany\\
2. Dept. of Neuroradiology, Klinikum rechts der Isar, Germany\\
3. Dept. of Computer Science and Engineering, Southern University of Science and Technology, China \\
4. Shenzhen Institude of Artificial Intelligence and Robotics for Society, China \\
5. Institude for Advanced Study, Technical University of Munich, Germany\\
\email{\{hongwei.li, bjoern.menze\}@tum.de}}
\maketitle              

\begin{abstract}
Synthesizing MR imaging sequences is highly relevant in clinical practice, as single sequences are often missing or are of poor quality (e.g. due to motion). Naturally, the idea arises that a target modality would benefit from multi-modal input, as proprietary information of individual modalities can be synergistic.
However, existing methods fail to scale up to multiple non-aligned imaging modalities, facing common drawbacks of complex imaging sequences. We propose a novel, scalable and multi-modal approach called \emph{DiamondGAN}. Our model is capable of performing flexible non-aligned cross-modality synthesis and data infill, when given multiple modalities or any of their arbitrary subsets, learning structured information in an end-to-end fashion.
We synthesize two MRI sequences with clinical relevance (i.e., double  inversion  recovery  (DIR)  and  contrast-enhanced  T1 (T1-c)), reconstructed from three common sequences. In addition, we perform a multi-rater visual evaluation experiment and find that trained radiologists are unable to distinguish synthetic DIR images from real ones.

\end{abstract}

\section{Introduction} \label{introduction}

In clinical practice, magnetic resonance imaging (MRI) datasets often consists of high-dimensional image volumes with multiple imaging protocols and repeated scans acquired at multiple time points.
Given the multiplicity of possible sequence parameters, protocols largely vary depends on the imaging centers, hindering their comparability. This often leads to repeated exams or severely limits the clinical information that can be drawn from those MRI studies. Particularly, in the case of multiple sclerosis, longitudinal comparisons of MRI studies are the main reason for treatment decisions and existing lesion quantification tools require complete identical modalities at multiple time points.
Potentially, cross-modality image synthesis technique can resolve those obstacles through efficient data infilling and re-synthesis.

Recently, generative adversarial networks (GANs) have been applied in translating MRI sequences, positron emission tomography (PET) and computed tomography (CT) images. Most of them are one-to-one cross-modality synthesis approaches, for example, PET \cite{wang20183d} synthesis and MRI sequences translation \cite{dar2019image}. A recent multi-modal synthesis method \cite{sharma2019missing} has limited scalability because the input and output modalities are required to be spatially aligned. Although there are several multi-domain translation algorithms \cite{choi2018stargan} in the computer vision community, these approaches design one-to-multiple domain translation but do not model the multiple-to-one domain mapping. Especially in medical images synthesis, \emph{multiple-to-one} cross-modality mapping is highly relevant as proprietary information of individual and non-aligned modalities can be synergistic.

There are three main challenges in the scenario of multi-modal cross-modality medical image synthesis: 1) the input and target modalities are assumed to be \emph{not} spatially-aligned because registration methods for aligning multiple modalities may fail, restricting the applicability of conventional regression approaches. 2) input modalities may be missing due to different clinical settings between centers, thus a traditional regression-based data infill would be restricted to the smallest uniform subset or rely on iterative data infill methods. 3) existing approaches have limited scalability, e.g. in a \emph{Cycle-GAN} \cite{zhu2017unpaired} setting, one would therefore have to train individual models for possible combinations of the input modalities.

\paragraph{Contributions}
1) We propose \emph{DiamondGAN}, which is a unified, scalable multi-modal generative adversarial network. It learns the multiple-to-one cross-modality mapping among non-aligned modalities using only a pair of generators and discriminators, optimized with a multi-modal cycle-consistency loss function.
2) We provide both qualitative and quantitative results on two clinically-relevant MRI sequences synthesis tasks, showing \emph{DiamondGAN's} superiority over baseline models.
3) We present the results of extensive visual evaluation, performed by fourteen experienced radiologists to confirm the quality of synthetic images.
\section{Methodology}
\subsection{Multi-Modal Cross-Modality Synthesis}
Given an input set of \emph{n} modalities: \emph{X = \{x$_{i}$$|$i = 1, ..., n\}} and a target modality \emph{T}. Our goal is to learn a generator \emph{G} that learns mappings from multiple input modalities to one target modality. We assume that 1) all the modalities, i.e., \emph{X} and \emph{T}, are \emph{not} spatially-aligned because it is rather difficult to obtain \emph{strictly} spatially-aligned images as mentioned in Section \ref{introduction}; 2) the input modalities can be any subset of \emph{X}, denoted as \emph{X'} during the training and inference stages as some modalities of a subject may be missing in clinical practice.

\begin{figure*}[t]
	\begin{center}
		\includegraphics[width=1\linewidth,height=0.32\linewidth]{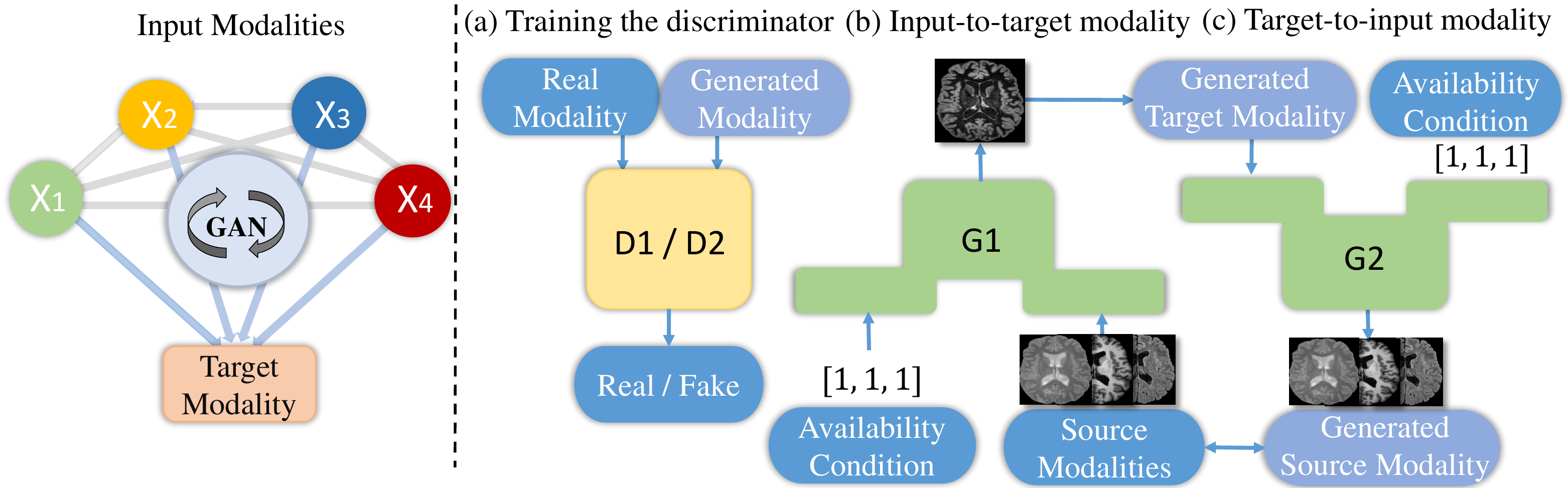}
	\end{center}
	\vspace{-0.45cm}
	\caption{Left: The high-level idea behind \emph{DiamondGAN}, which is capable of learning mappings between any subset of multiple input modalities (\emph{X}) to a target modality in a single model. This mapping represents a diamond-shape topology. Right: Overview of \emph{DiamondGAN}. It consists of two modules, a pair of discriminators \emph{D} and a pair of generators \emph{G}. (a) \emph{D1} and \emph{D2} learn to distinguish between the real and synthetic images from multi-modal input and the target output respectively. (b) \emph{G1} takes both multi-modal input and the condition as input and generates a target modality. The condition \emph{c} is a binary vector: $c = \{c_{1}, c_{2}, ..., c_{n}\}$, where $c_{i}$ indicates the corresponding input modality as available (1) or not (0). It is spatially replicated and concatenated with the input modalities in the feature level. (c) \emph{G2} tries to generate the original modalities from the synthetic target modality given the original availability condition.}
	\label{fig:DiamondGAN} \vspace{-0.4cm}
\end{figure*}

We enforce \emph{G} to be capable of translating any subset \emph{X'} into a target modality \emph{T} using a condition \emph{c} which indicates the presence of the input modalities, i.e., \emph{G(X', c) $\rightarrow$ T}. This condition handles the missing modality issue and makes it a scalable model in both the training and the inference stages. We further introduce a multi-modal cycle-consistency loss to handle the "non-aligned modalities" issue among the input and output. Fig. \ref{fig:DiamondGAN} illustrates the main idea of our proposed approach.
We regularly generate the condition \emph{c} and the corresponding multi-modal data \emph{X$_{c}$} of all possible combinations, so that \emph{G} learns to flexibly translate the arbitrary multi-modal input. As mentioned in the caption of Fig.~\ref{fig:DiamondGAN}, we use an \emph{availability} condition to serve as an indicator of the input modalities. It is spatially replicated to the image size ($1 \times H \times W$) and is a part of the two-stream network input. In the case of 3 modalities as the input, the condition $c = [1, 1, 1]$ would indicate that every input modality is given.
\vspace{-0.2cm}

\vspace{-0.3cm}
\subsubsection{Multi-Modal Reconstruction Loss} \label{multi_modal_loss}
We aim to train \emph{G} to guarantee that a generated target modality preserves the content of its input modalities.
The input modalities are assumed to be not spatially aligned or not from the same subject as mentioned above. In this situation, the traditional cycle loss \cite{zhu2017unpaired} as well as the regression loss \cite{isola2017image} would fail to tackle the multi-modal and non-alignment issues.
To alleviate the two problems, we extend the traditional cycle-consistency loss \cite{zhu2017unpaired} to a multi-modal one. Specifically, we concatenate the source modalities into a multi-channel input and define a multi-channel output as the target modality. We then simultaneously train two generators $G_{1}:X \rightarrow T$ and $G_{2}:T \rightarrow X$ in a cycle-consistency fashion. Please note that the output target modality is in multiple channels which correspond to the input modalities. The loss function of the generator is defined as:
\vspace{-0.25cm}
\begin{equation}
	\mathcal{L}_{rec} = \mathbb{E}_{X,T, c}[||X-G_{2}(G_{1}(X, c), c)||_{1} + ||T-G_{1}(G_{2}(T, c), c)||_{1}]
	\end{equation}
\vspace{-0.25cm}	
\subsubsection{Adversarial Loss}
To make the generated images indistinguishable from real images, we adopt an adversarial loss:
\begin{equation} \label{equation_1}
\begin{aligned}
\mathcal{L}_{adv} = {}&\mathbb{E}_{X, T}\{log~[D_{1}(X)\cdot D_{2}(T)]\} \\
                        + &\mathbb{E}_{X,T,c}\{log~[(1 - D_{2}(G_{1}(X,c)))\cdot (1 - D_{1}(G_{2}(T,c)))]\}
\end{aligned}
\end{equation}
where \emph{G$_{1}$} generates a target modality \emph{G$_{1}$(X, c)} conditioned on the presence of input modalities \emph{X}, while \emph{D$_{1}$} tries to distinguish between real input modalities and generated ones. Similarly, \emph{G$_{2}$} generates the original input modalities \emph{G$_{2}$(T, c)} conditioned on the presence of original input modalities \emph{X} and \emph{D$_{2}$} tries to distinguish between the real target modality and the generated one. The generators try to minimize this objective, while the discriminators to maximize it.
\vspace{-0.3 cm}
\subsubsection{Full Objective}
The objective functions to optimize \emph{D} and \emph{G} respectively are
\begin{equation}
	\mathcal{L}_{D} = - \mathcal{L}_{adv};~~\mathcal{L}_{G} = \mathcal{L}_{adv} + \lambda_{rec}\mathcal{L}_{rec}
\vspace{-0.1cm}
\end{equation}
where $\lambda_{rec}$ is the hyper-parameter that balances the reconstruction loss and adversarial loss.

\subsection{Implementation}

\subsubsection{Two-Stream Network Architecture}
To leverage the information from both input modalities and corresponding availability conditions, we build a two-stream network architecture based on the popular encoder-decoder network \cite{johnson2016perceptual}. It takes the multi-modal images and condition as two inputs and merges them in the feature level. This network contains stride-2 convolutions, residual blocks \cite{he2016deep} and fractionally-strided convolutions (1/2 stride). We use 6 blocks for the input size of $N \times H \times W$, where $N$, $H$ and $W$ are the number of modalities, height and width of the images respectively. The input and availability conditions pass through two encoders and are merged in the last feature layer before the decoder. \emph{PatchGANs} \cite{johnson2016perceptual} is used for the discriminator network, which classifies the patch feature maps to real or fake, instead of using a fully-connected layer.
\vspace{-0.3 cm}
\subsubsection{Training Details}
We apply two recent techniques to stabilize the training of the model. First, for $\mathcal{L}_{adv}$ (Eq. \ref{equation_1}), we replace the negative log likelihood objective by a least-squares loss \cite{mao2017least}.
Second, to reduce the model oscillation, we update the discriminators using a history of generated images rather than the ones produced by the latest generators, as proposed in \cite{shrivastava2017learning}. Thus we put the 25 previously generated images in an image buffer. We set $\lambda_{rec}$ = 10 in Equation 3 for all the experiments. We use the Adam solver \cite{kingma2014adam} with a batch size of 5. All networks were trained from scratch with a learning rate of 0.0002 and for 20 epochs. When given $n$ input modalities, for each epoch the parameters in both generator and discriminator are updated for \emph{2$^{n}$-1} times given \emph{2$^{n}$-1} training subsets of input modalities excluding empty set. The implementations of our model are available in \url{https://github.com/hongweilibran/DiamondGAN}. 

\vspace{-0.3cm}
\subsection{Visual Rating and Evaluation Protocol}
Quantitative evaluation of generated images in terms of standard scores for errors and correlation remains a debatable task \cite{borji2019pros}. Additionally, the evaluation with common metrics such as PSNR and MAE \cite{welander2018generative} would not tell us to whether the algorithm captures clinically relevant small substructures.
Therefore, we strive to get experts' estimates of the image quality. We design a multi-rater quality evaluation experiment.
Neuro-radiologists rated the images in a browser-application.
In each trial, they were provided with two images. On the left side, one real source image of a T1 or Flair images is presented. On the other side, a paired image of the target modality is shown which is either a real image or a generated one. The displayed paired images were randomly chosen in the pool of generated images and real ones. This particular setup enables the experts to identify very small inconsistency or implausibility between the two images immediately. For evaluation, the experts were asked to rate the plausibility of the image on the right based on the real image on the left, to assign a 6-star rating, where 6 stars denoted a perfectly plausible image and 1 star a completely implausible image. The images were presented in 280 trials.
\section{Experiments}
\subsubsection{Datasets}
\emph{Dataset 1} consists of 65 scans of patients with MS lesions from a local hospital, acquired with a multi-parametric protocol, which includes co-registered Flair, T1, T2, double inversion recovery (DIR) and contrast-enhanced T1 (T1-c) after skull-stripping. The first three modalities are common modalities in most MS lesion exams. DIR is a MRI pulse sequence, which suppresses signal from the cerebrospinal fluid and the white matter, enhancing the inflammatory lesion. T1-c is a MRI sequence which requires a paramagnetic contrast agent (usually gadolinium) that reduces the T1 relaxation time and thereby increases the signal intensity. Synthesizing DIR and T1-c is of clinical relevance because it can substantially reduce medical costs. We mainly report our result on \emph{Dataset 1}. 
Additional \emph{Dataset 2} is used to demonstrate that our approach can work on multiple datasets with incomplete and non-aligned modalities. It is a part of the public MICCAI-WMH dataset \cite{kuijf2019standardized}, and includes 40 subjects with two modalities (Flair and T1). 
2D axial slices are used for training the network. All the slices are cropped or padded to a uniform size of 240 $\times$ 240 and intensity values are rescaled to [-1, 1].
\vspace{-0.3 cm}
\subsubsection{Reconstructing DIR and T1-c from Common Modalities}
We perform two image synthesis tasks on two clinically-relevant MRI sequences (DIR and T1-c), using three common modalities (i.e., Flair, T1 and T2).
We separate the \emph{Dataset 1} into a training set, a validation set and a test set, resulting in 30 scans (2015 slices for each modality) for training and 35 scans for testing (2100 slices for each modality). To obtain the optimal hyper-parameters of the model, we use 5 out of the 30 training scans as a validation set.
A common approach for quantitative evaluation of medical GAN images is to calculate relative errors and signal to noise ratio between the synthetic image and the real image \cite{welander2018generative}. Table \ref{table:structured_results} shows the results
of peak signal-to-noise ratio (PSNR)
and mean absolute error (MAE) by comparing the synthetic images and real T1-c and DIR images. For the synthetic DIR and T1-c images, we report the highest PSNR and the lowest MAE for a combined T1+T2+Flair input to our model. In the DIR synthesis experiment, the listed scores of using multiple inputs to our GAN are comparable (MAE 0.058-0.065). Whereas, the scores for single inputs are substantially worse (MAE 0.073-0.084). For the T1-c synthesis task, we find that any combination of multi-modal inputs involving the T1 modality (MAE 0.045-0.048) results in better scores compared to other inputs. This indicates that our model successfully extracts the relevant information, as T1-c is a T1 scan with a contrast enhancing agent. For comparison, we implement \emph{CycleGAN} \cite{zhu2017unpaired} to perform one-to-one cross-modality synthesis, the best results of \emph{CycleGAN} are listed in Table. \ref{table:structured_results}. For DIR synthesis, using Flair images as the input of \emph{CycleGAN} achieves the highest PSNR and lowest MAE while for T1-c, using T1 as the input gets the best performance. The proposed model outperforms \emph{CycleGAN} in both tasks. We further replace a part of the training Flair and T1 images in \emph{Dataset 1} with images from \emph{Dataset 2} (totally 794 images for each modality) and we find the result on same testing set is comparable to using the original \emph{Dataset 1}.

Wilconxon signed-rank tests are conducted on the PSNR and MAE pairs generated by \emph{DiamondGAN} (with 3 modalities) and \emph{CycleGAN} respectively. Although the improvements of PSNR and MAE look small in whole image level, they are statistically significant (p-value$<$0.0001) in the case of DIR in Table \ref{table:structured_results}. This improvement is highly relevant for biomaker synthesis and for pathological evaluation especially in the case of MS lesions with small volumes.
\vspace{-0.15cm}
\begin{figure*}[t]
    \vspace{-0.25cm}
	\begin{center}
		\includegraphics[width=1.1\linewidth,height=0.5\linewidth]{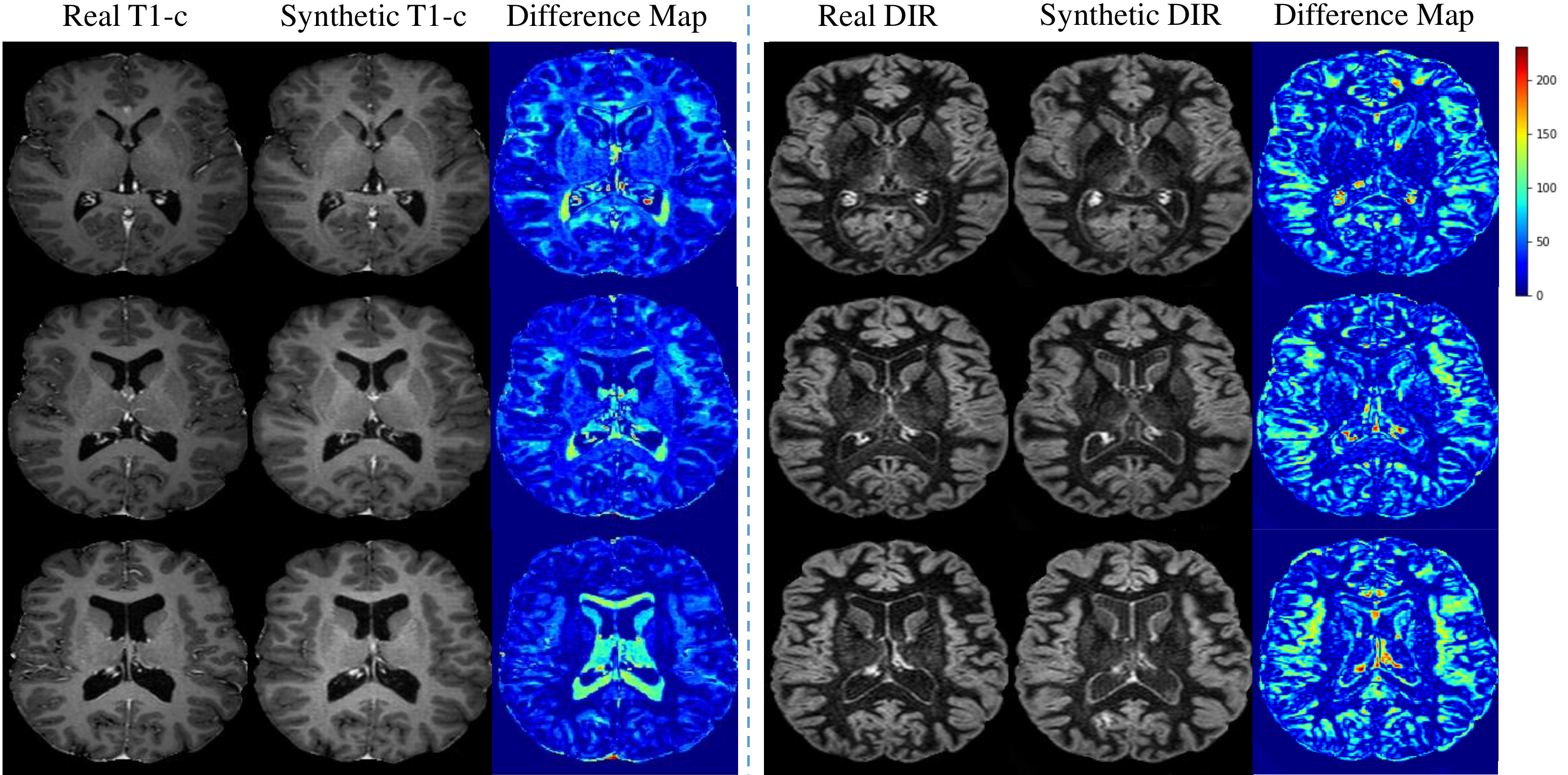}
	\end{center}
	\vspace{-0.45cm}
    	\caption{Samples of synthetic T1-c and DIR images given the combination of T1, T2 and Flair modalities. Difference images are generated and visualized in heat maps. The synthetic images preserve the tissue contrast and the anatomy information. However, we find more differences in synthetic DIR images than in synthetic T1-c ones, especially around the brain boundary. This could be due to the alignment error by registration methods.}
	\label{fig:samples} \vspace{-0.45cm}
\end{figure*}

\makeatletter
\def\thickhline{%
  \noalign{\ifnum0=`}\fi\hrule \@height \thickarrayrulewidth \futurelet
   \reserved@a\@xthickhline}
\def\@xthickhline{\ifx\reserved@a\thickhline
               \vskip\doublerulesep
               \vskip-\thickarrayrulewidth
             \fi
      \ifnum0=`{\fi}}
\makeatother

\newlength{\thickarrayrulewidth}
\setlength{\thickarrayrulewidth}{2\arrayrulewidth}

\begin{table}
[t]
\vspace{-0.2cm}
\newcommand{\tabincell}[2]{\begin{tabular}{@{}#1@{}}#2\end{tabular}}
\renewcommand\arraystretch{1}
\centering
\vspace{-0.2cm}
\caption{Quantitative evaluation of our generated images compared to the real DIR and T1-c image using PSNR and MAE as evaluation metrics. Results show that the generated images benefit from a multi-modal input. $\uparrow$ indicates that higher values corresponds to better image qualities.}\label{table:structured_results}.
\vspace{-0.2cm}  
\begin{tabular}{lcccc}
\thickhline
            &~~\textbf{DIR $_{PSNR}$$\uparrow$}~~ & ~~\textbf{DIR $_{MAE}$$\downarrow$}~~ & ~~\textbf{T1-c $_{PSNR}$$\uparrow$}~~& ~~\textbf{T1-c $_{MAE}$$\downarrow$}
    \\
    \thickhline
\emph{CycleGAN} \cite{zhu2017unpaired} & 17.34 & 0.068   & 20.36 &   0.045\\
\thickhline
$DiamonGAN_{T1}$     &15.46  & 0.084   &20.21  & 0.048  \\
$DiamonGAN_{T2}$          & 15.99          & 0.073          & 19.34          & 0.054           \\
$DiamonGAN_{Flair}$     & 16.16          & 0.078          & 17.15          & 0.068           \\
$DiamonGAN_{T1+T2}$      & 17.41          & 0.065          & 20.75          & 0.046           \\
$DiamonGAN_{T2+Flair}$   & 18.58          & 0.059          & 19.78          & 0.051           \\
$DiamonGAN_{T1+Flair}$    & 18.02          & 0.062          & 20.40          & 0.047           \\
$DiamonGAN_{T1+T2+Flair}$  & \textbf{18.63} & \textbf{0.058} & \textbf{20.86} & \textbf{0.045}  \\
\hline
\end{tabular}
\end{table}
\vspace{-0.1cm}

\begin{figure*}[t]
	\begin{center}
        \vspace{-0.1cm}
		\includegraphics[width=1\linewidth]{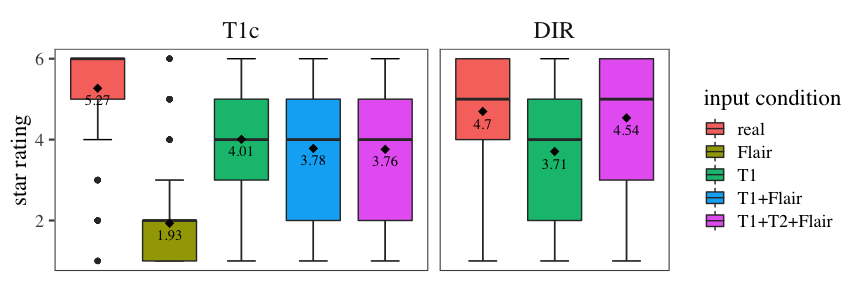}
	\end{center}
    	\caption{Box plots showing the rating scores of synthetic images and real ones for T1-c modality on the left and DIR modality on the right. The means are shown as black numbers. \emph{DiamondGAN} achieves comparable plausibility levels for the DIR modality.}
	\label{fig:boxplot}
\end{figure*}
\vspace{-0.3cm}
\subsubsection{Visual Evaluations by Neuroradiologists}
Fourteen neuro-radiologists with median 5+ years of professional experience participated. Each evaluated 210 synthetic images and 70 original images. The 210 synthetic images are generated enforcing 6 different input conditions in which each condition includes 35 samples. The rating results of the 14 raters are averaged and the box plots of the results are shown in Figure \ref{fig:boxplot}. For the synthesis of T1-c images, we found that three multi-modal combinations (i.e., \emph{T1}, \emph{T1+Flair} and \emph{T1+T2+Flair}) gave comparable results, while the ones based solely on a Flair were consistently rated as implausible. 
The plausibility of DIR images synthesized with $T1+T2+Flair$ input was rated on average 0.83 stars higher than that with solely T1 input. This is plausible as the DIR is a complex sequence containing proprietary information, its synthesis thus benefits from multiple input sources. For the synthetic images with \emph{T1+T2+Flair} input, the experts assigned an identical rating to the synthetic and original images (4.54 stars \emph{vs} 4.7 stars).

We conduct Wilcoxon rank-sum tests on the paired rating scores of synthetic and real images from 14 raters on 6 conditions which results in 6 pairs of 14 observations. Results show that the pair of rating scores on synthetic DIR images by \emph{T1+T2+Flair} input and real DIR images are not significantly different (p-value = 0.1432) while all other pairs are significantly different (p-values $<$ 0.0001). This demonstrates that trained radiologists are unable to distinguish our synthetic DIR images from real ones. Furthermore, the experts ratings for the individual conditions of synthetic images are in agreement with the metrical evaluation in Table \ref{table:structured_results}. For T1-c synthesis, the PSNR and MAE scores are consistently good when T1 modality is fed to \emph{DiamondGAN}. 
\section{Conclusion and Discussion}
This work introduces a novel approach for multi-modal medical image synthesis, with extensive multi-rater experiments and statistical tests.
This multi-modal approach allows us to mine the structured information inside the existing extensive MRI sequences.
Pathological evaluation is the ultimate goal of this work. Our approach is evaluated by clinical partners who contributed the datasets. We compared synthetic DIR sequence with conventional FLAIR sequence in a MS lesions detection task in a cohort study. The proposed \emph{DiamondGAN} has the potential to reduce medical costs in clinical practice.

\subsubsection{Acknowledgement}
This work is suppport by Technische Universit\"at M\"unchen - Institute for Advanced Study, funded by the German Excellence Initiative and European Union 7$^{th}$ Framework Programme under grant agreement No. 291763. HL and BW are supported by the funding from Zentrum Digitalisierung Bayern.

{
	\bibliographystyle{splncs03}
	\bibliography{egbib}
}

\end{document}